\documentclass[10pt,twocolumn,letterpaper]{article}

\usepackage{iccv}
\usepackage{times}
\usepackage{epsfig}
\usepackage{graphicx}
\usepackage{amsmath}
\usepackage{amssymb}
\usepackage{multirow}
\usepackage{makecell}
\usepackage{array}
\usepackage{booktabs}
\usepackage{float} 
\usepackage{pifont}
\usepackage{subcaption}
\RequirePackage[format=plain,labelformat=simple,labelsep=period,font=small,compatibility=false]{caption}
\usepackage[pagebackref=true,breaklinks=true,letterpaper=true,colorlinks,bookmarks=false]{hyperref}
\usepackage[capitalize]{cleveref}
\crefname{section}{Sec.}{Secs.}
\Crefname{section}{Section}{Sections}
\Crefname{table}{Table}{Tables}
\crefname{table}{Tab.}{Tabs.}
\iccvfinalcopy % *** Uncomment this line for the final submission

\def\iccvPaperID{11069} % *** Enter the ICCV Paper ID here

\ificcvfinal\fi

\begin{document}
%%%%%%%%% TITLE
\title{DWRSeg: Rethinking Efficient Acquisition of Multi-scale Contextual \\Information for Real-time Semantic Segmentation}

\author{Haoran Wei, Xu Liu, Shouchun Xu, Zhongjian Dai, Yaping Dai, Xiangyang Xu\thanks{Corresponding author.}\\
Beijing Institute of Technology\\
No. 5, South Street, Zhongguancun, Haidian District, Beijing\\
{\tt\small $\{$hrwei, liuxu, 3220200814, daizhj, daiyaping, xxy1970$\}$@bit.edu.cn}
}

\maketitle
\ificcvfinal\thispagestyle{empty}\fi
%%%%%%%%% ABSTRACT
\begin{abstract}
   Many current works directly adopt multi-rate depth-wise dilated convolutions to capture multi-scale contextual information simultaneously from one input feature map, thus improving the feature extraction efficiency for real-time semantic segmentation.
   However, this design may lead to difficult access to multi-scale contextual information because of the unreasonable structure and hyperparameters. 
   To lower the difficulty of drawing multi-scale contextual information, we propose a highly efficient multi-scale feature extraction method, which decomposes the original single-step method into two steps, Region Residualization-Semantic Residualization.
   In this method, the multi-rate depth-wise dilated convolutions take a simpler role in feature extraction: performing simple semantic-based morphological filtering with one desired receptive field in the second step based on each concise feature map of region form provided by the first step, to improve their efficiency.
   Moreover, the dilation rates and the capacity of dilated convolutions for each network stage are elaborated to fully utilize all the feature maps of region form that can be achieved.
   Accordingly, we design a novel Dilation-wise Residual (DWR) module and a Simple Inverted Residual (SIR) module for the high and low level network, respectively, and form a powerful DWR Segmentation (DWRSeg) network. Extensive experiments on the Cityscapes and CamVid datasets demonstrate the effectiveness of our method by achieving a \textbf{state-of-the-art} trade-off between accuracy and inference speed, in addition to being lighter weight. Without pretraining or resorting to any training trick, we achieve an mIoU of 72.7\% on the Cityscapes test set at a speed of 319.5 FPS on one NVIDIA GeForce GTX 1080 Ti card, which exceeds the latest methods of a speed of 69.5 FPS and 0.8\% mIoU. The code and trained models are publicly available. 
\end{abstract}

\begin{figure}[t]
  \centering
   \includegraphics[width=1.0\linewidth]{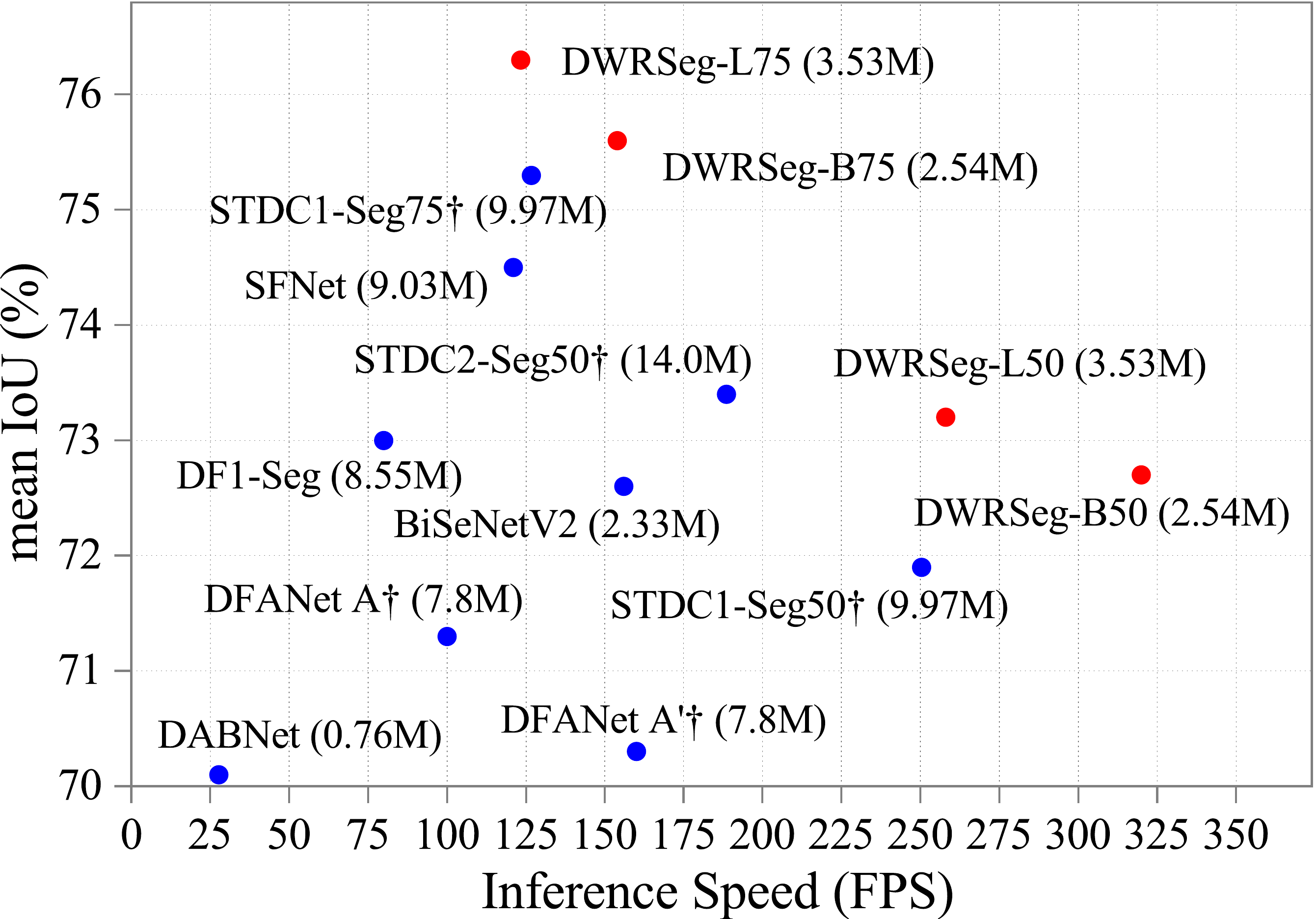}
\caption{Speed-accuracy performance comparison on the Cityscapes test set. Our methods are presented as red dots, while the other methods are presented as blue dots. $\dagger$ represents the methods pretrained on ImageNet. The parameters of the methods are shown in parentheses. Our approaches achieve a \textbf{state-of-the-art} speed-accuracy trade-off, in addition to being lighter weight.}
   \label{fig:miou}
\end{figure}

%%%%%%%%% BODY TEXT
\section{Introduction}

Semantic segmentation plays a significant role in many fields, including automatic driving, robot sensing, satellite remote sensing and medical imaging. To meet the real-time requirements in real scenes and reduce the application expense, research on lightweight and efficient methods is of great significance.

Efficient feature extraction is a core issue in real-time semantic segmentation tasks. In many current works, such as ESPNet (V2) ~\cite{mehta2018espnet, mehta2019espnetv2}, DABNet ~\cite{li2019dabnet} and CGNet~\cite{wu2020cgnet}, modules capturing multi-scale contextual information from one input feature map based on multi-rate depth-wise dilated convolutions are designed to improve the efficiency of feature extraction for real-time semantic segmentation. 
% However, these designs all have fatal defects, which lead to their inability to effectively elicit multi-scale contextual information. 
% The specific phenomenon is that a large number of weights in depth-wise dilated convolutions are rarely learned in these designs, especially for those with larger dilation rates. 
However, these designs all have fatal defects, leading a large number of weights in depth-wise dilated convolutions to be rarely learned, especially for those with larger dilation rates, so that multi-scale contextual information cannot be effectively elicited. 
We believe that the causes are as follows:
(1) Directly applying dilated depth-wise convolutions with multiple receptive fields simultaneously on each feature map 
may invalidate some receptive fields because not all the receptive fields are required by a feature map with a certain representation.
(2) Input feature maps with complex representations make it more difficult for depth-wise large-rate dilated convolutions to establish long-distance semantic connections. PReLU is commonly used in all these methods, leading to more complicated feature presentations.
(3) There is a lack of reasonable design for the receptive fields in different network stages. For instance, in ESPNet (V2), the same large-scale receptive field is used at the low stage of the network as at the high stages, resulting in the decline in feature extraction efficiency at the low stage. ESPNet, CGNet and DABNet all tend to utilize massive receptive fields to acquire more contextual information, far exceeding the receptive field size requirements.
For the above reasons, it is difficult for these methods to capture multi-scale contextual information from one complex feature map using multiple depth-wise dilated convolutions with undesirable dilation rates. Our aim is to capture multi-scale information more easily and efficiently.

\begin{figure}[t]
  \centering
   \includegraphics[width=1.0\linewidth]{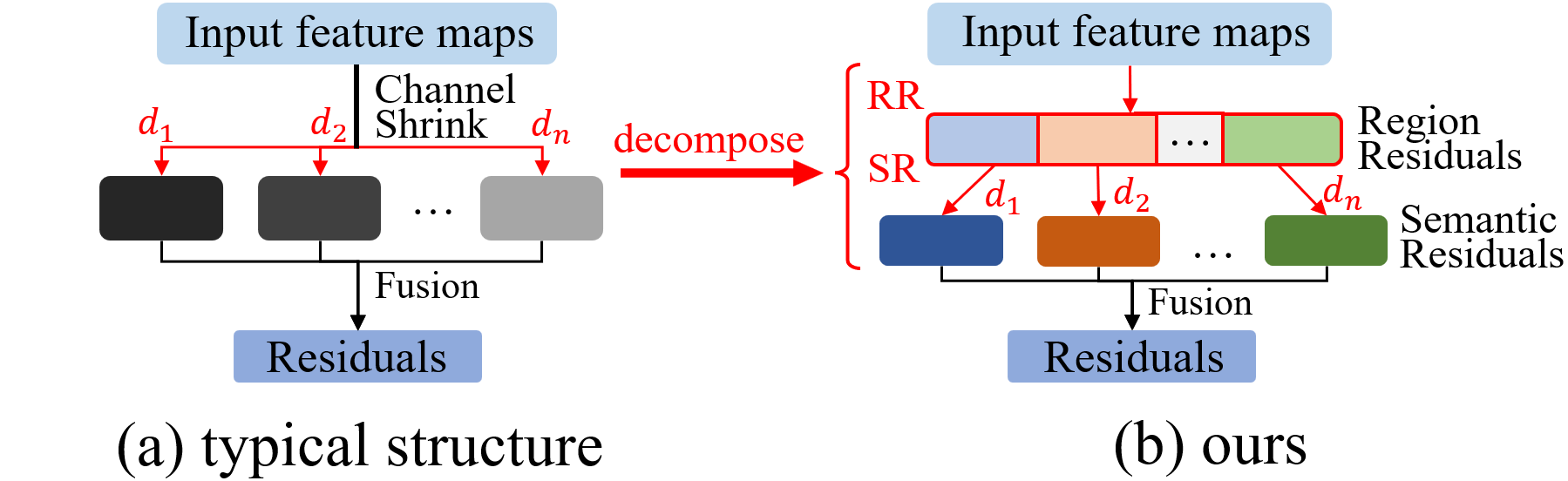}
\caption{(a) presents the typical structure for drawing multi-scale contextual information and (b) presents our structure, in which RR and SR denote Region Residualization and Semantic Residualization respectively, $d_x,x \in (1,n)$ denotes dilated depth-wise convolution with a dilation rate $d_x$.}
   \label{fig:intro}
\end{figure}

To this end, we propose a highly efficient two-step method to drawing multi-scale contextual information for real-time semantic segmentation.
In this method, the previous single-step method is decomposed into two steps for the purpose of lowering the difficulty of drawing multi-scale contextual information, as shown in \cref{fig:intro}.
In the first step, concerned features with concise regional expressions of different sizes are generated, which is referred to as \emph{Region Residualization}. In the second step, only one depth-wise dilated convolution with a desired receptive field is used for semantic-based morphological filtering on each concise feature map of region form, which is referred to as \emph{Semantic Residualization}.
By this means, the role of the multi-rate depth-wise dilated convolutions in feature extraction is transformed from a difficult one to an easy one: from obtaining as much complex contextual information from complexly expressed feature maps as possible to performing simple morphological filtering on each concisely expressed feature map with a desired dilation rate.
The concise feature maps of region form clear and simplify the learning goal of the dilated depth-wise convolutions and make the learning process more orderly.
A single applied dilation rate can avoid redundant receptive fields that are not required by the feature map with a certain region expression.
Moreover, different feature materials can be adaptively assigned in the first step according to the size of the receptive field in the second step to reversely match the receptive field. 
Furthermore, to fully utilize all feature maps with different region sizes that can be achieved in each network stage, it is necessary to elaborate the dilation rates and the capacity of dilated convolutions to match the different receptive field requirements in each network stage.

% To this end, we propose a highly efficient two-step feature extraction method for the purpose of lowering the difficulty of drawing multi-scale contextual information. In this method, we make great following efforts: (1) Transform the role of depth-wise dilated convolutions: from obtaining more complex semantic information from complexly expressed feature maps as much as possible to performing simple morphological filtering on concisely expressed feature maps. (2) Adaptively apply the corresponding receptive field on each feature map, which requires specific receptive field for morphological filtering. Therefore, the two-step method is designed as follows:
%  \emph{First step}, concise concerned features taking regions of different sizes are generated, which is referred to as \emph{Region Residualization};
%  \emph{Second step}, a dilated convolution with a specific dilation rate is used for morphological filtering on each concise regional feature map, which is referred to as \emph{Semantic Residualization}.
%  Thus, different feature materials can be adaptively assigned in the first step according to the size of the receptive field in the second step, so as to reversely match the receptive field. 
%  Moreover, to fully utilize all feature maps with different region sizes that can be achieved in each stage, it is necessary to elaborate the dilation rates and the capacity of dilated convolutions to match different receptive field requirements in each network stage.
 
Accordingly, we propose a novel Dilation-wise Residual (DWR) module that adopts the designed two-step feature extraction methods in the high stages of the network. A Simple Inverted Residual (SIR) module is specifically designed for the low stages, which is adjusted from the DWR module to meet the small receptive field requirement. Based on these modules, a DWRSeg network is established, which greatly promotes network performance in real-time semantic segmentation.

Our main contributions can be summarized as follows:

1.	We delve into the mechanism of multi-scale feature extraction with depth-wise dilated convolutions and design a highly efficient two-step residual feature extraction method (Region Residualization – Semantic Residualization) to greatly improve the efficiency of capturing multi-scale information in real-time semantic segmentation.

2.	We propose a novel DWR module and SIR module with elaborate receptive field sizes for the upper and lower stages of the network based on our proposed method, and a powerful DWRSeg network is formed based on these modules for real-time semantic segmentation.

3.	We conduct extensive experiments to verify the effectiveness of our methods, and our architecture achieves impressive results on Cityscapes, as shown in \Cref{fig:miou}. Specifically, we obtain an mIoU of 72.7\% on the Cityscapes test set at a speed of 319.5 FPS on one NVIDIA GeForce GTX 1080 Ti card, which exceeds the latest methods of a speed of 69.5 FPS and an mIoU of 0.8\%.

%------------------------------------------------------------------------
\section{Related Work}

\subsection{Real-time Semantic Segmentation}

Real-time semantic segmentation networks can be divided into 3 categories according to the adopted backbones~\cite{mo2022review}. (\romannumeral1) \emph{lightweight classification model-based method.} The DFANet~\cite{li2019dfanet} utilizes a modified Xception~\cite{chollet2017xception} model as the backbone and aggregates features in a cascaded manner.  SFNet~\cite{li2020semantic} adopts the ResNet~\cite{he2016deep}, ShuffleNet V2~\cite{ma2018shufflenet} and DF~\cite{li2019partial} series as the backbone in the encoder and proposes an FAM to learn semantic flow in the decoder. BiSeNetV1~\cite{yu2018bisenet} uses Xception39~\cite{chollet2017xception} as the backbone of the context path and designs a feature fusion module (FFM) to merge features. (\romannumeral2) \emph{multibranch architecture-based method.} ICNet~\cite{zhao2018icnet} takes cascaded images into multiresolution branches and utilizes a cascade feature fusion unit to quickly carry out high-quality segmentation. BiSeNetV2~\cite{yu2021bisenet} uses two branches to obtain semantic and detail information; then, it fuses both types of feature representations through a guided aggregation layer. (\romannumeral3) \emph{specialized backbone-based method.} ENet~\cite{paszke2016enet} employs a special initial block and several bottleneck blocks to make up the backbone. The first two blocks heavily reduce the input size to avoid expensive computation.
% DABNet~\cite{li2019dabnet} utilizes a novel depth-wise asymmetric bottleneck (DAB) module to construct the backbone, which is able to create a sufficient receptive field and densely utilize the contextual information.
\subsection{Receptive Field}

The design of the receptive field is of great importance for real-time semantic segmentation networks since it is related to both speed and accuracy. 
\cite{luo2016understanding} showed that the effective receptive field only
accounts for a fraction of the full theoretical receptive field. DeepLab V2~\cite{chen2017deeplab} uses an atrous spatial pyramid pooling (ASPP) module composed of parallel convolutions with different receptive fields to extract different features, thus capturing image context information at multiple scales. Following DeepLab V2, many algorithms~\cite{chen2017rethinking,chen2018encoder,zhao2017pyramid,yang2018denseaspp} have achieved advanced performance by using ASPP-like modules at the bottom of the backbone. Different from the above, some works~\cite{wang2018understanding,wu2020cgnet,yuan2020multi} explored designing multiscale receptive field modules and interspersing them in a specialized backbone. In this paper, we propose an efficient lightweight backbone to extract features with a scalable receptive field.

%-------------------------------------------------------------------------
\section{Proposed Method}
In this section, we first introduce the overall architecture of our proposed network – DWRSeg for real-time semantic segmentation. Then we present the design details of each component of our network.
\subsection{Overall architecture}
\begin{figure*}
  \centering
    \includegraphics[width=1.00\linewidth]{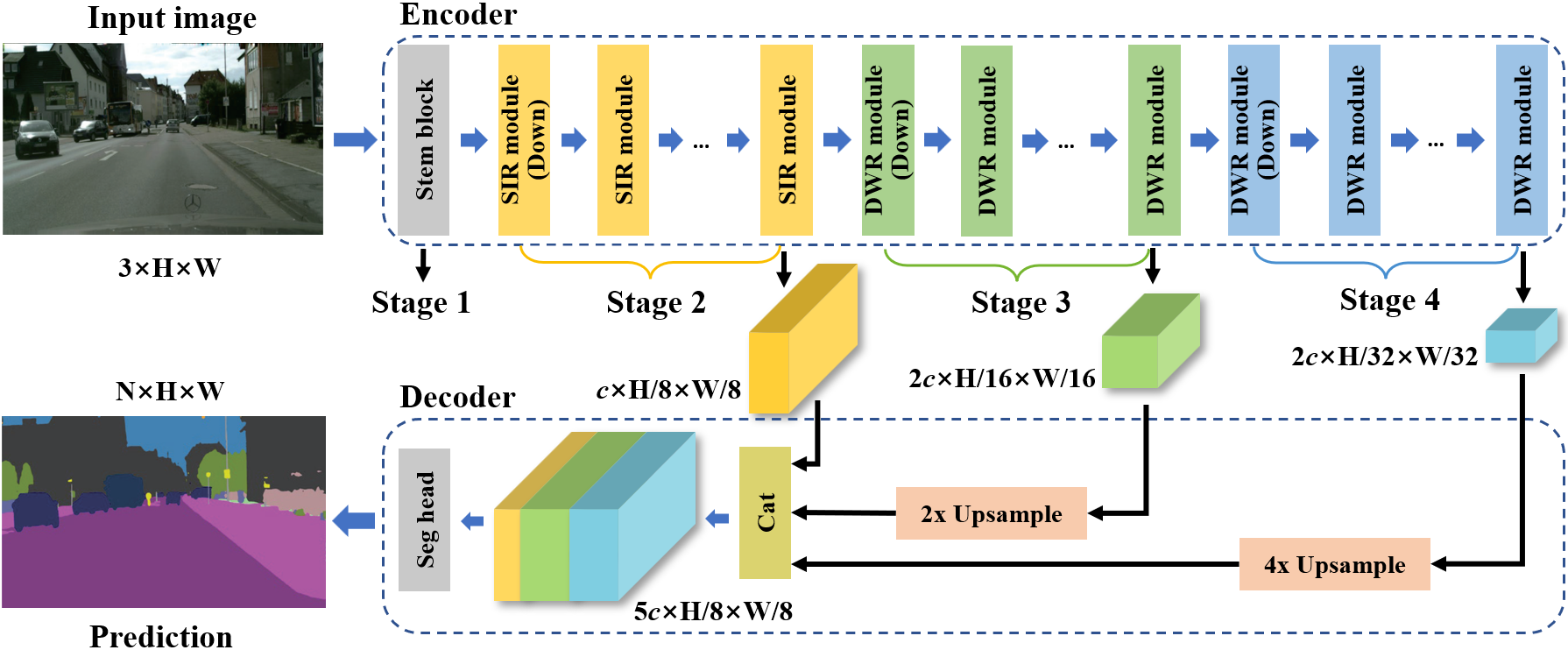}
\caption{Illustration of the entire network structure. The structure is relatively simple; it is a typical encoder-decoder setup. The encoder includes four stages, a stem block, SIR modules, DWR modules (2 branches) and DWR modules (3 branches). The stem block is used for initial processing, the SIR modules are used to extract features from the low stage, and the DWR modules are used to extract features from the higher stages. The output of the upper three layers enters the decoder, which is a simple FCN-like style. Feature maps from the higher stages are first upsampled and then concatenated with feature maps from the lower stage. Finally, a SegHead is used to process the concatenated feature maps to conduct the final prediction. H and W denote the input image size, N denotes the number of classes, and $c$ denotes the base number of the feature map channels.}
  \label{fig:st}
\end{figure*}

The whole architecture is shown in \Cref{fig:st}, which is highly simple but efficient. In general, our proposed network is an encoder-decoder setup. The encoder consists of four stages: stem block, the low stage of the SIR modules and two high stages of the DWR modules. A DWR module is used for feature extraction in the high stages with a \emph{two-step} multi-scale contextual information acquisition method. An SIR module, which is modified from the DWR module to meet a smaller receptive field requirement, is used in the low stage. A stem block is used for the initial processing of the input images. The output feature maps of the upper three stages enter the decoder, which is in a simple FCN~\cite{long2015fully}-like style. The final prediction is generated from the decoder and directly supervised by the ground truth. No auxiliary supervision is needed during the training progress. We refer to our network as DWRSeg according to the core module, i.e., the DWR module.

We carefully tune the hyperparameters in the whole network until the best trade-off between accuracy and efficiency is reached. Finally, we report two versions: DWRSeg-Base (DWRSeg-B) and DWRSeg-Large (DWRSeg-L). \Cref{tab:detail} shows the detailed structure of our DWRSeg networks.

\begin{table}

\renewcommand{\arraystretch}{0.8}
\centering
\begin{tabular}{cccccc}
\toprule[1.5pt]
\multirow{2}{*}{Stages} & \multirow{2}{*}{R
%\begin{tabular}[c]{@{}c@{}}Output\\Ratio\end{tabular}
} &
\multirow{2}{*}{C}&
\multirow{2}{*}{B}&

\multicolumn{2}{c}{Repetitions} \\ \cline{5-6} 
        &  &   &  & 
        DWRSeg-B\rule{0pt}{10pt} & DWRSeg-L\\
\midrule[1pt]
Image   & 1  & 3         &           &          &           \\\midrule
\begin{tabular}[c]{@{}c@{}}Stem\\block\end{tabular}  & 1/4   & 64   & &   &\\\midrule
\begin{tabular}[c]{@{}c@{}}SIR\\modules\end{tabular}              & 1/8                                                                 & 64                   & 1                    & 7                   & 8                   \\ \midrule
\begin{tabular}[c]{@{}c@{}}DWR\\modules\end{tabular}             & 1/16                                                                  & 128                  & 2                    & 3                  & 8                    \\ \midrule
\begin{tabular}[c]{@{}c@{}}DWR\\modules\end{tabular}             & 1/32                                                                  & 128                  & 3                    & 3                  & 3                    \\\midrule
\begin{tabular}[c]{@{}c@{}}Decoder\\cat\end{tabular}              & 1/8                                                                 & 320                  &                      &                   &                      \\\midrule
\begin{tabular}[c]{@{}c@{}}Decoder\\conv1\end{tabular}            & 1/8                                                                 & 128                  &                      &                   &                      \\\midrule
\begin{tabular}[c]{@{}c@{}}Decoder\\conv2\end{tabular}           & 1/8                                                                & N                    &                      &                   &                      \\\midrule[1pt]
Flops   &   &&   & 13.62G      &16.42G    \\\midrule
Params     &      &       &    & 2.54M   &        3.53M \\
\bottomrule[1.5pt]
\end{tabular}
\caption{Detailed DWRSeg architecture. R, C and B denote the ratio between the output size and input image size, output channels and branch number of the module respectively. Flops are estimated for an input resolution of 3$\times$512$\times$1024, and N denotes the number of classes.}
\label{tab:detail}
\end{table}

\subsection{Dilation-wise Residual Module}
The core of our work lies in the design of the DWR module with an efficient \emph{two-step} multi-scale contextual information acquisition method.

\begin{figure}[t]
  \centering
  \begin{subfigure}{0.613\linewidth}
    \includegraphics[width=1.\linewidth]{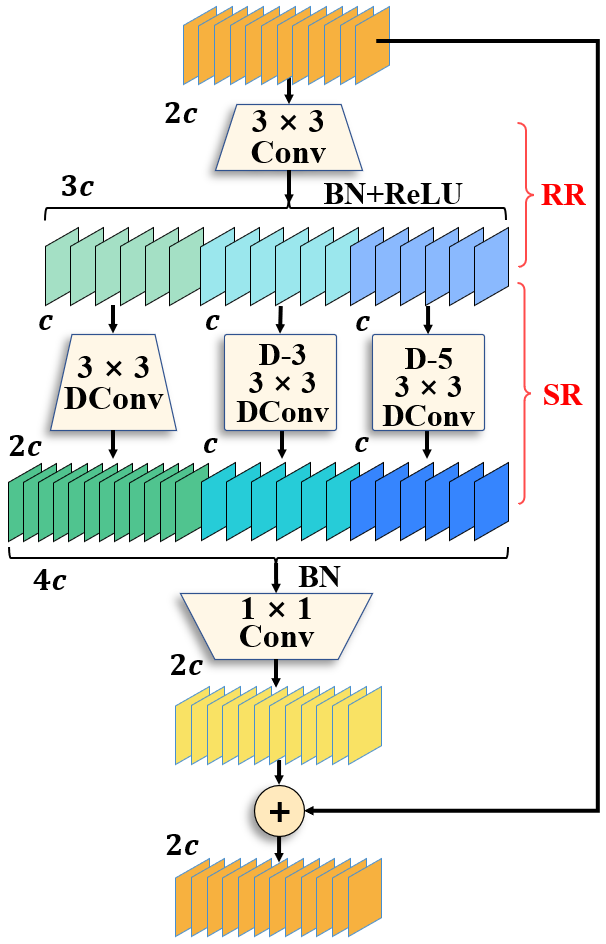}
    \caption{DWR module}
    \label{fig:dwrmodule}
  \end{subfigure}
  \hfill
  \begin{subfigure}{0.377\linewidth}
    \includegraphics[width=1\linewidth]{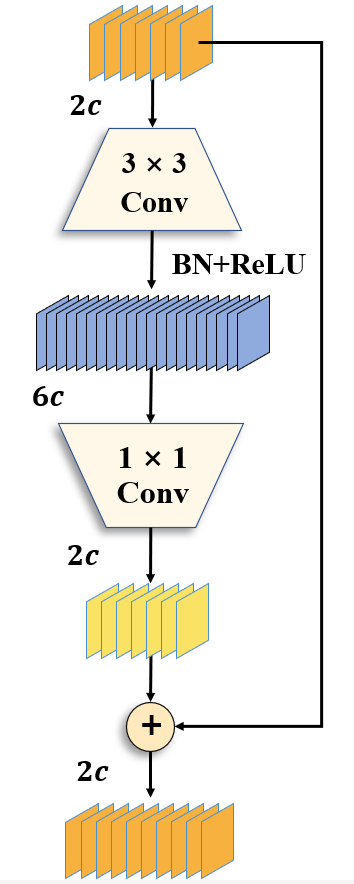}
    \caption{SIR module}
    \label{fig:sirmodule}
  \end{subfigure}
\caption{Illustration of the structure of a 3-branch DWR module and SIR module. (a) and (b) present a three-branch DWR module of the structure for the high-level network and an SIR module of the single-path structure for the low-level network, respectively. RR and SR denote Region Residualization and Semantic Residualization, respectively. Conv denotes convolution, DConv denotes depth-wise convolution, D-$n$ denotes dilated convolution with a dilation rate of $n$, + in a circle denotes the addition operation, and $c$ denotes the base number of feature map channels.}
   \label{fig:dwr}
\end{figure}

\textbf{Design idea and structure.}
Generally, the DWR module is designed in a residual fashion, as shown in \Cref{fig:dwrmodule}. Inside the residual, a two-step method is used to draw multi-scale contextual information efficiently, and then feature maps generated with multi-scale receptive fields are fused.
Specifically, the previous single-step multi-scale contextual information acquisition method is decomposed into a \emph{two-step} method to reduce the acquisition difficulty. 

\begin{figure}[t]
  \centering
    \includegraphics[width=1\linewidth]{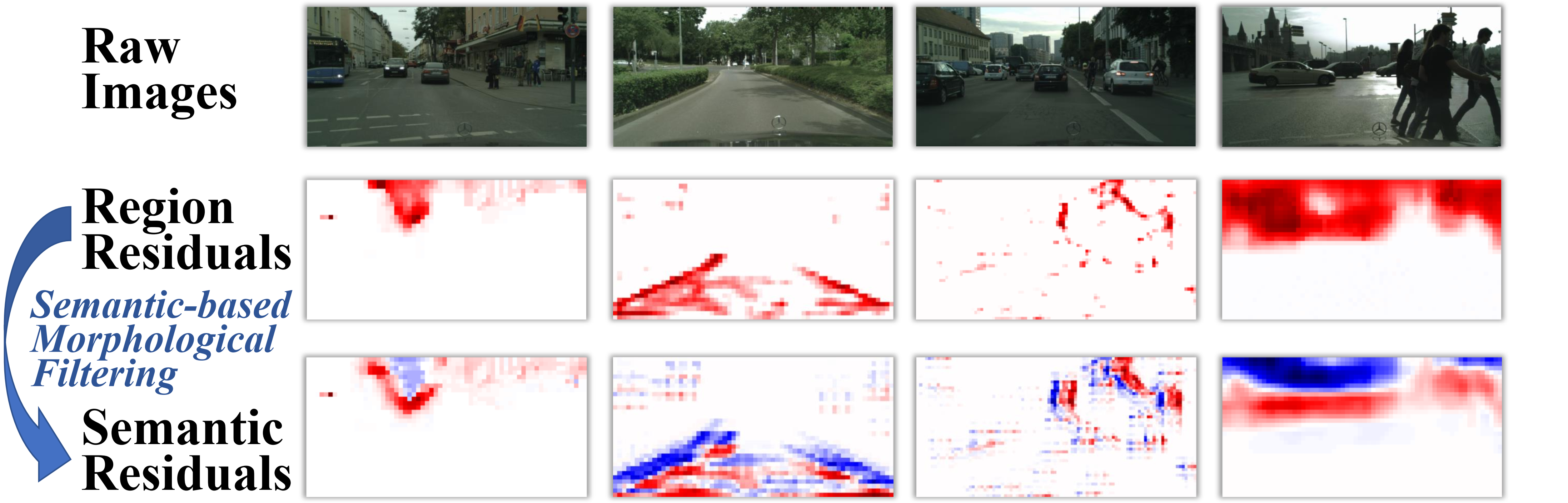}
\caption{Illustration of the visualized heatmaps of feature maps from Region Residualization and Semantic Residualization outputs. The upper row shows the raw images. The middle row shows the feature maps obtained by Region Residualization. The red, blue and white colors represent positive, negative and no features respectively.
Some spatial regions of different sizes can be activated, and the feature maps are concise. The lower row shows the corresponding feature maps obtained by Semantic Residualization. Simple semantic-based morphological filtering is performed on the regional features with the desired receptive fields.}
   \label{fig:rrsr}
\end{figure}

\emph{Step \uppercase\expandafter{\romannumeral1}}. Concerned residual features are generated from input features, which is referred to as Region Residualization. In this step, a series of concise feature maps with region form of different sizes are generated, as shown in \Cref{fig:rrsr}, to serve as materials for morphological filtering in {Step  \uppercase\expandafter{\romannumeral2}}.
This step is implemented by a common 3x3 convolution combined with a batch normalization (BN) layer and an ReLU layer. The 3x3 convolution is used for preliminary feature extraction.
The ReLU activation function, rather than a commonly used PReLU layer\cite{he2015delving}, plays an important role of activating regional features and making them concise. 

\emph{Step \uppercase\expandafter{\romannumeral2}}. Multi-rate dilated depth-wise convolutions are adopted separately to morphologically filter regional features of different sizes, which is referred to as Semantic Residualization. The results of morphological filtering are shown in \Cref{fig:rrsr}.
Only one single desired receptive field is applied to each channel features to prevent possible redundant receptive fields. In reality, the required concise regional feature maps can be wisely learned in the first step according to the size of the receptive field in the second step, to reversely match the receptive field. 
To implement this step, the regional feature maps are first divided into sereral groups, and then dilated depth-wise convolutions with different rates are conducted on the different groups. 

Through the two-step Region Residualization-Semantic Residualization method, the role of multi-rate depth-wise dilated convolutions is changed from laboriously obtaining as much complex semantic information as possible to simply performing morphological filtering with one desired receptive field on each concisely expressed feature map. 
The feature maps in region form make the role of depth-wise dilated convolutions simplified to morphological filtering, thus ordering the learning process.
Therefore, multi-scale contextual information can be secured more efficiently.

After drawing multi-scale contextual information, the multiple outputs are aggregated. Specifically, all the feature maps are concatenated. Next, BN is performed on the maps. Then, a point-wise convolution is adopted to merge features, forming the final residuals. Finally, the final residuals are added to the input feature maps to construct a stronger and more comprehensive feature representation.

\textbf{Parameter design.}
It is necessary to elaborate the dilation rates and the capacity of the dilated convolutions to fully utilize all feature maps with different region sizes that can be achieved in each network stage.
The feature maps of different region sizes generated by Region Residualization are not only related to the corresponding dilation rates in Semantic Residualization, but are also limited by the receptive fields required at the current stage.

As the network level increases, the acceptance of larger receptive fields also increases since semantic enhancement allows convolutions to establish longer-term spatial links. However, excessively large receptive fields, such as dilation rates exceeding 7, always have little effect.
Accordingly, in network stage 4, 3 branches of dilated convolution with dilation rates of 1, 3, and 5 are set, while in stage 3, the third branch is abandoned to squeeze the receptive field to avoid invalid computation. 

Moreover, as it is always more difficult for convolutions to establish connections over a larger spatial span directly, and a long-span connection needs the assistance of small-span connections, a small receptive field is always important in each stage. Accordingly, the output channel capacity of the first branch is expanded to be twice that of other branches.

% $\textbf{Design details.}
% Although the structure of the module is uncomplicated, some design details are worth noting. 
% (1) To fully meet the requirements of all receptive fields capacity, the convolution in Region Residualization extends the channel to draw more regional residual materials.
% Therefore, this module has an inverted bottleneck structure, and in stage 4, the expansion ratio is set to 1.5.
% (2) We believe that different sizes of receptive fields are required differently. 
% In addition, as it is more difficult for convolutions to establish connections over larger spatial span directly, a small dilation rate is always important in each stage. 
% Accordingly, in stage 4, 3 branches of dilated convolution with dilation rates of 1, 3, and 5 are set, and the ratio of output channel capacity of each branch is set to 2:1:1. 
% (3) We believe that the receptive field size is required diminishedly as the stages descend, since the link distances that can be established by convolutions become shorter as semantics weaken.
% As a result, the DWR modules in stage 3 abandon the third branch to squeeze the receptive field.
% (4) The nonlinearity in the DWR module is relatively low compared with that in other residual blocks (such as the CG block~\cite{wu2020cgnet}, the DAB module~\cite{li2019dabnet} and the GE layer~\cite{yu2021bisenet}). Excessive nonli$nearity may not yield better results, but it may lead to inefficiency.

\subsection{Simple Inverted Residual Module}
The Simple Inverted Residual (SIR) module is designed from the DWR module to meet the requirement for a smaller receptive field size in the lower stage, to maintain high feature extraction efficiency. The structure is shown in \Cref{fig:sirmodule}. Two modifications are made based on the DWR module: (1) The multi-branch dilated convolution structure is removed and only the first branch is retained to compress the receptive field. (2) The 3x3 depth-wise convolution (Semantic Residualization) is removed due to its minimal contribution. Because of the large size and weak semantics of the input feature maps, too little information can be collected by a single-channel convolution. Therefore, one-step feature extraction is more efficient than two-step feature extraction in the low stage.
As a result, quite a simple module that inherits the residual and inverted-bottleneck structures of the DWR module is finally obtained.

\subsection{Other components}
\begin{figure}[t]
  \centering
  \begin{subfigure}{0.603\linewidth}
  \centering
    \includegraphics[width=0.98\linewidth]{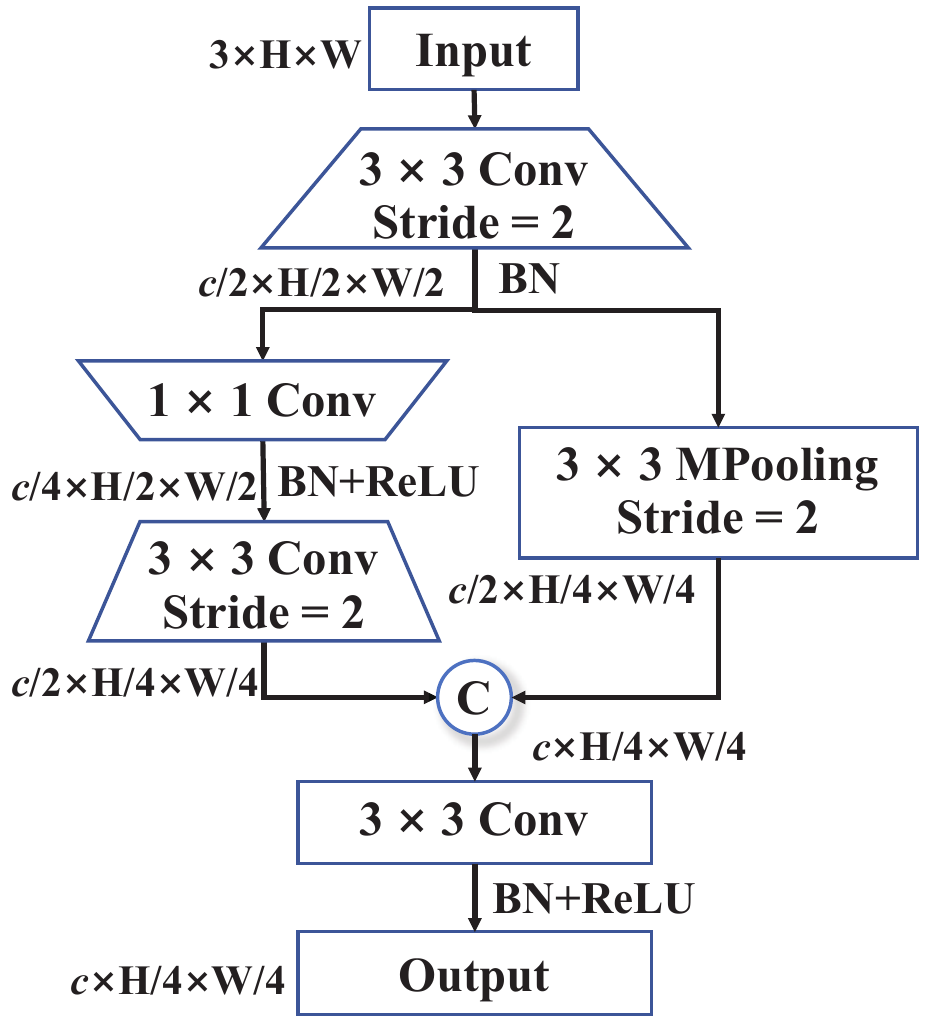}
    \caption{Stem block}
    \label{fig:stemblock}
  \end{subfigure}
  \hfill
  \begin{subfigure}{0.386\linewidth}
  \centering
    \includegraphics[width=0.68\linewidth]{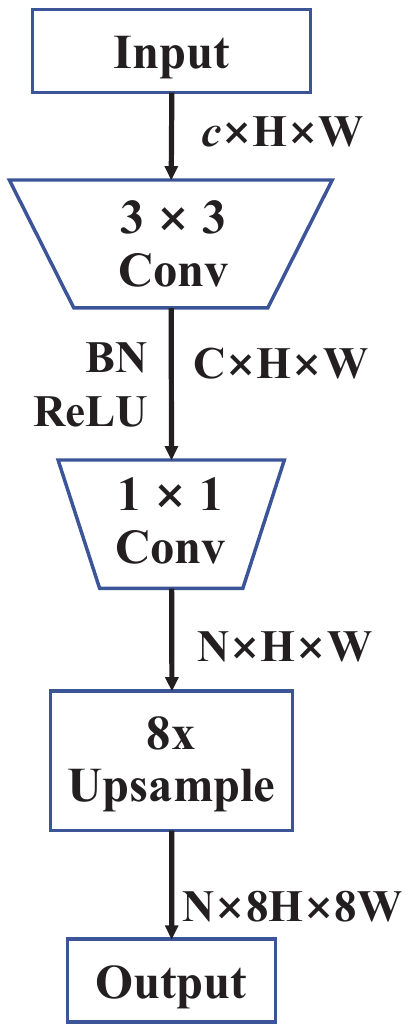}
    \caption{Segmentation head}
    \label{fig:sh}
  \end{subfigure}
\caption{Illustration of the structure of the stem block and SegHead. (a) presents the structure of the stem block, which is used to process the input images; (b) presents the structure of the SegHead, which is used to convert feature maps into the final prediction. Conv denotes convolution, Mpooling denotes max pooling, C in a circle denotes the concatenate operation, $c$ denotes the base number of feature map channels, and H and W denote the input size.}
   \label{fig:SIRandstem}
\end{figure}

\textbf{Stem block.}
Following BiSeNetV2~\cite{yu2021bisenet}, a stem block is adopted as the first stage of our network, as shown in \Cref{fig:stemblock}. Two adjustments are made to the stem block presented in BiSeNetV2. (1) We remove the activation layer for the first convolution to retain more valid information from the original images. (2) The last convolution does not perform feature shrinkage. This structure has low computational cost and effective feature expression ability.

\textbf{Decoder}
The decoder adopts a simple FCN-like structure, as shown in \Cref{fig:st}. The output feature maps of stages 3 and 4 are first upsampled and then concatenated with feature maps from stage 2. After that, a BN layer acts on the concatenated feature maps. Finally, a segmentation head (SegHead) is used for prediction. The SegHead structure is illustrated in \Cref{fig:sh}. A 3×3 convolution with a BN layer and an ReLU layer is set to merge the feature maps. Afterwards, a point-wise convolution is used for prediction. The feature maps are upsampled to the input size in the end. The hyperparameters in the decoder are all shown in \Cref{tab:detail}.

%------------------------------------------------------------------------
\section{Experiments}
In this section, we first introduce the experimental settings. Next, we investigate the desired receptive fields and the effects of each component of our proposed approach on the Cityscapes test set. Finally, we report our accuracy and speed results on different benchmarks compared with those of other algorithms.
\subsection{Experimental Settings}
\textbf{Datasets.}\ Cityscapes~\cite{cordts2016cityscapes} is an urban street scenes semantic segmentation dataset. Only 19 classes are used for experiments.
The Cambridge-driving Labeled Video Database (CamVid)~\cite{brostow2008segmentation} is a road scene dataset. Only 11 classes are involved in our experiments for a fair comparison with other methods.

\textbf{Implementation Details.} For training, we use mini-batch stochastic gradient descent (SGD) with 0.9 momentum and 0.0005 weight decay to train our model. For all datasets, the batch size is set as 16. As a common configuration, we utilize a 'poly' learning rate policy with the power set to 0.9 and the initial learning rate set as 0.02. In addition, we train the models for 1,000 and 600 epochs for the Cityscapes and CamVid datasets, respectively. We adopt cross-entry loss with Online Hard Example Mining~\cite{shrivastava2016training} to optimize the learning task.
Data augmentation includes color jittering, random horizontal flipping, random cropping and random resampling. The resize scale range is [0.25, 1.5], and the cropped resolution is 1280$\times$640 for training Cityscapes. For CamVid, the resize scale range is [0.5, 1.5], and the cropped resolution is 960 $\times$ 720.
All our networks are trained \textbf{without pretraining on ImageNet}.

For inference, we do not adopt any evaluation tricks, e.g., sliding-window evaluation and multiscale testing, which can improve accuracy but are time consuming. For Cityscapes, we first resize images to 1024$\times$512 or 1536$\times$768 to perform the inference and then resize the prediction to the original image size.

We conduct all our experiments based on PyTorch-1.10, CUDA 11.3, CUDNN 8.2.0 and TensorRT 8.2.3.0. Two NVIDIA RTX 3090 GPUs are used for training, and the measurement of inference time is executed on an NVIDIA GTX 1080Ti GPU with a batch size of 1 for benchmarking.

\begin{figure*}
  \centering
  \begin{minipage}[b]{0.22\linewidth}
  \centering
    \includegraphics[width=1.0\linewidth]{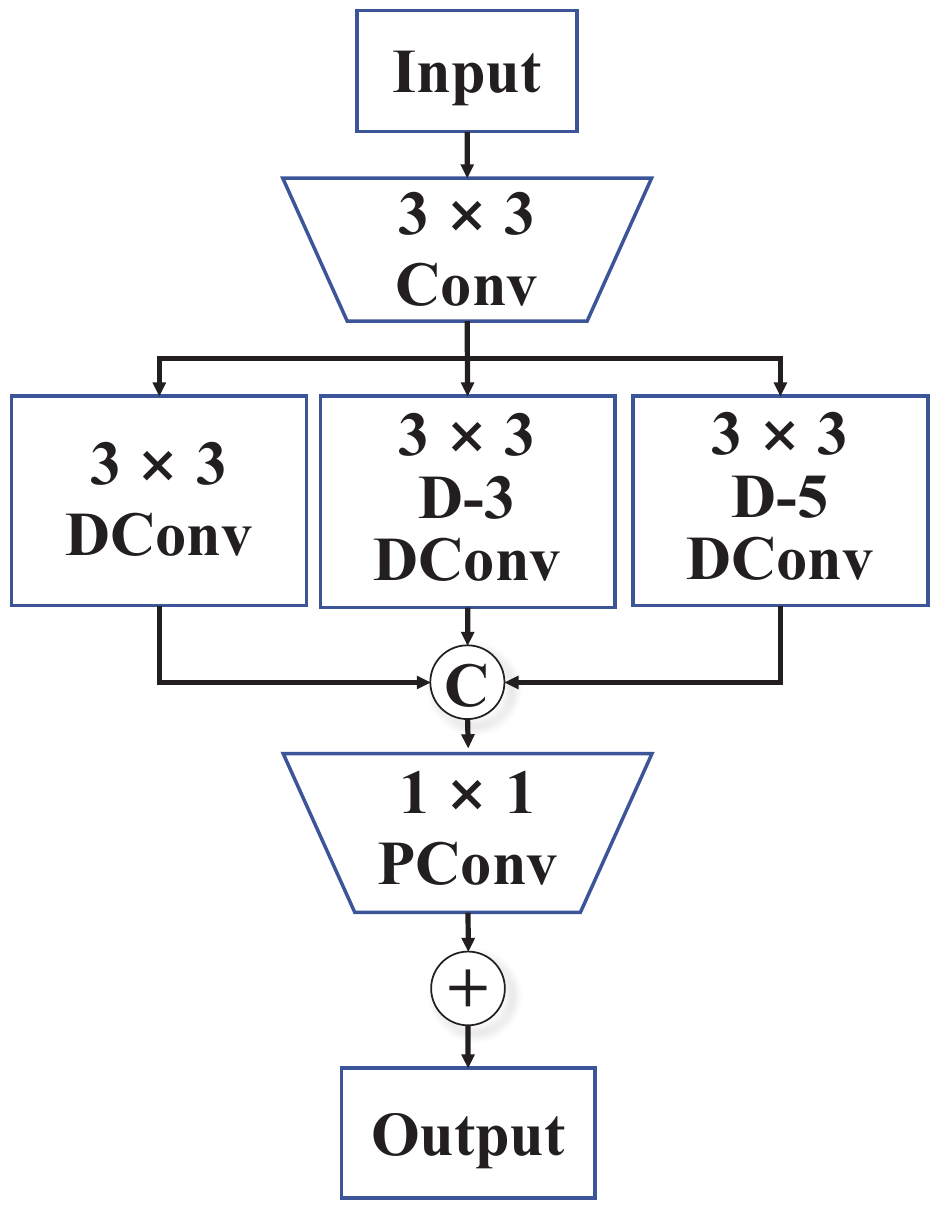}
\caption{The designed module to test the demand for receptive fields, whose each feature map receives multiple receptive fields.}
   \label{fig:idea}
  \end{minipage}
  \hfill
  \begin{minipage}[b]{0.76\linewidth}
  \centering
  \begin{subfigure}{0.325\linewidth}
    \includegraphics[width=0.99\linewidth]{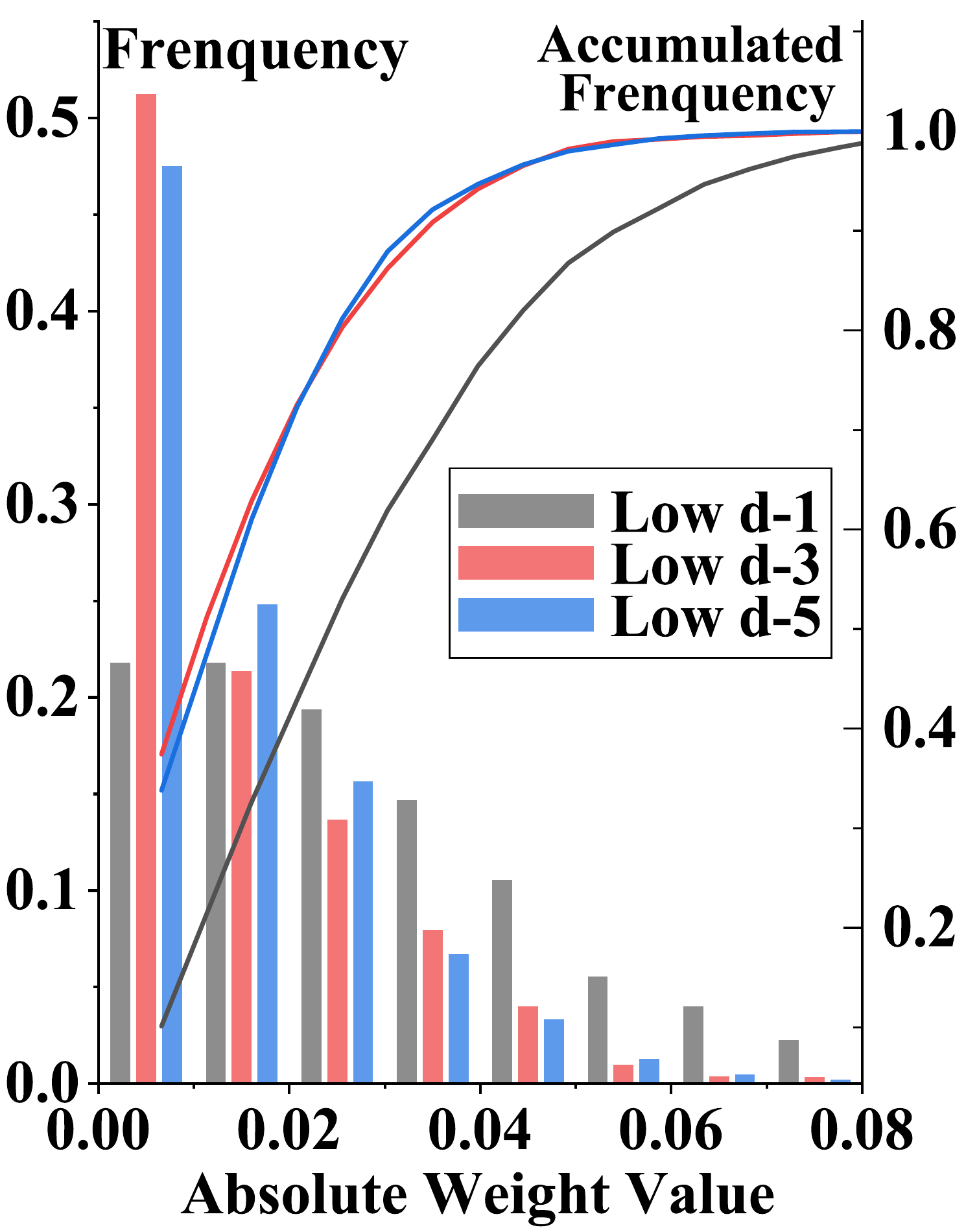}
    \caption{Low Stage}
    \label{fig:low}
  \end{subfigure}
  \hfill
  \begin{subfigure}{0.325\linewidth}
    \includegraphics[width=0.99\linewidth]{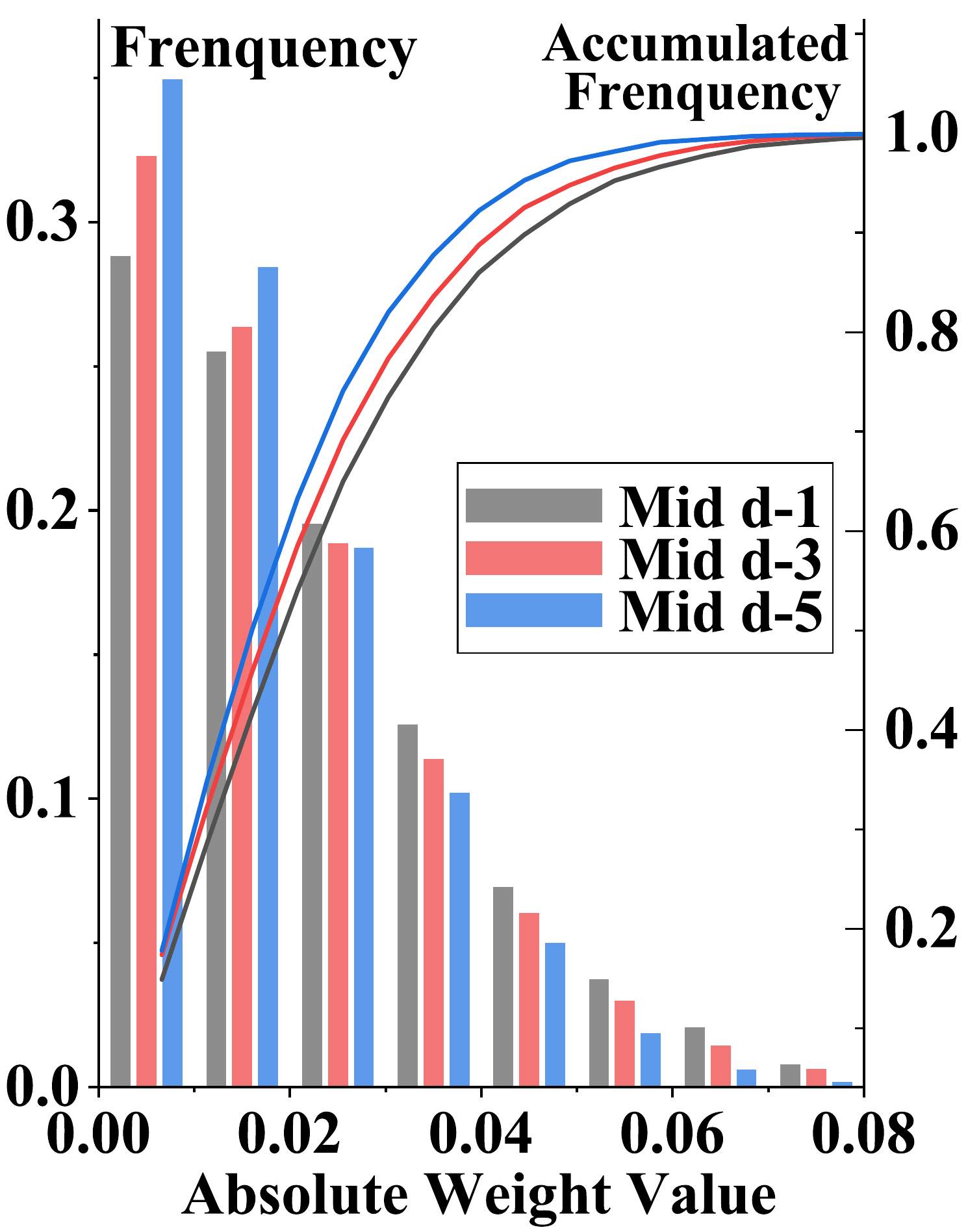}
    \caption{Medium Stage}
    \label{fig:mid}
  \end{subfigure}
  \hfill
  \begin{subfigure}{0.325\linewidth}
    \includegraphics[width=0.99\linewidth]{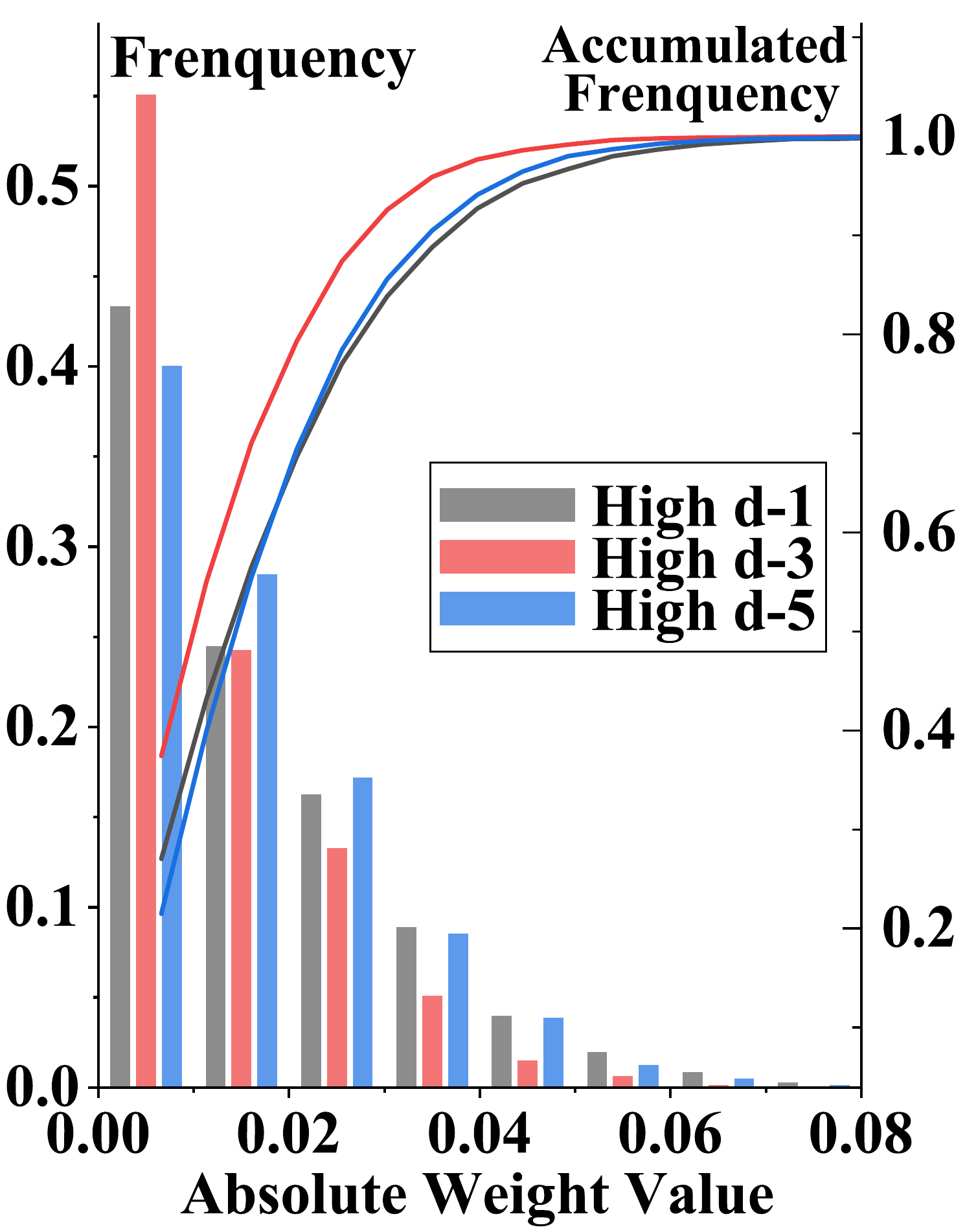}
    \caption{High Stage}
    \label{fig:high}
  \end{subfigure}
\caption{Illustration of the probability mass function (PMF) and cumulative distribution function (CDF) of the absolute weights in point-wise convolution corresponding to each branch of the top three stages. (a), (b) and (c) present the PMF and CDF of the low, medium and high stages, respectively. The histograms show the PMF, and the curves show the CDF. Gray, blue and red represent the absolute weights corresponding to the branches with dilation rates of 1, 3 and 5, respectively.}

  \label{fig:PMFCDF}
\end{minipage}
\end{figure*}

\subsection{Experiments on Desired Receptive Fields}
We design an experiment to verify our design idea for the dilation rates and the capacity of the depth-wise dilated convolutions in each stage and each module of the network. Our design is based on the view that the acceptance of large receptive fields increases with semantics enhance. However, excessively large receptive fields are completely ineffective. Moreover, regardless of the network stage, a small receptive field retains importance.

A special module, whose structure is shown in \Cref{fig:idea}, is built for this experiment. In this module, several dilated depth-wise convolutions with different rates work on one regional feature map so that all semantic-residualizations with different receptive field sizes compete for the regional feature map. We replace all DWR modules and SIR modules in DWRSeg-L with this module and then observe the weights in point-wise convolution corresponding to each branch; we use this to determine the importance of different receptive field sizes for the current stage. 

The probability mass function (PMF) and the cumulative distribution function (CDF) of the weights of each branch in different stages are shown in \Cref{fig:PMFCDF}. The convolution weights responsible for features from the small receptive field in the low stage are much greater than the others. In the middle stage, the weights for the larger receptive field increase considerably, while those for the small receptive field remain the greatest. In regard to the high stage, the weights for the largest receptive field are almost the same as those for the small. 
Therefore, it is verified that the demand for a wider receptive field grows from the low level to the high level. However, even at the topmost stage, the demand for a small receptive field size does not decrease significantly. Consequently, this experiment verifies that our designed dilation rates and capacity are desirable.

\subsection{Ablation Studies}
This section introduces the ablation experiments to validate the effectiveness of each component in our method. To eliminate the influence of different model sizes, except for the last ablation experiment, the number of modules in all models was adjusted to a similar inference speed.

\begin{table}[]
\renewcommand{\arraystretch}{0.85}
    \centering
    \begin{tabular}{m{0.6cm}<{\centering}|m{0.8cm}<{\centering} m{1.9cm}<{\centering} 
    m{1.4cm}<{\centering}
    m{1.3cm}<{\centering}
    }
    \toprule[1.5pt]
    Expt.
    &\multicolumn{2}{c}{Method} & mIoU(\%) & FPS\\ \midrule[1pt]
    \multirowcell{2}{\hypertarget{1}{\uppercase\expandafter{\romannumeral1}}}&\multirowcell{2}{
    % Experiment\hypertarget{1}{
    SS
    %\uppercase\expandafter{\romannumeral1}}
    } &w (B/L) & 76.3 & 123.4 \\
    &&w/o (MS)& 75.6 & 121.3 \\
    \midrule[1pt]
    \multirowcell{3}{\hypertarget{2}{\uppercase\expandafter{\romannumeral2}}}&\multirowcell{3}{
    % Experiment \hypertarget{2}{
    $\alpha$
    %\uppercase\expandafter{\romannumeral2}}
    }&2:1:1 (B/L) & 76.3 & 123.4 \\
    &&1:1:1 & 76.0 & 121.8 \\
    &&3:1:1 & 75.6 & 123.4 \\
    \midrule[1pt]
    \multirowcell{2}{\hypertarget{3}{\uppercase\expandafter{\romannumeral3}}}&\multirowcell{2}{
    % Experiment \hypertarget{3}{
    $\beta$
    % \uppercase\expandafter{\romannumeral3}}
    }&1.5 (B/L) & 76.3 & 123.4 \\
    &&3 & 75.3 & 122.3 \\
    \midrule[1pt]
    \multirowcell{6}{\hypertarget4{\uppercase\expandafter{\romannumeral4}}}&\multirowcell{6}{
    % Experiment \hypertarget{4}{
    NL
    % \uppercase\expandafter{\romannumeral4}
    }
    &B/L & 76.3 & 123.4 \\
    &&Sitch 1 & 75.2 & 124.3 \\
    &&Sitch 2 & 75.0 & 127.8 \\
    &&Sitch 3 & 75.6 & 126.5 \\
    &&Sitch 4 & 76.2 & 120.3 \\
    &&Sitch 5 & 76.3 & 118.5 \\
    \midrule[1pt]
    \multirowcell{2}{\hypertarget{5}{\uppercase\expandafter{\romannumeral5}}}&\multirowcell{2}{
    % Experiment \hypertarget{5}{
    PWC
    % \uppercase\expandafter{\romannumeral5}}
    }&w/o (B/L) & 76.3 & 123.4 \\
    &&w& 75.6 & 122.6 \\
    \midrule[1pt]
    \multirowcell{4}{\hypertarget{6}{\uppercase\expandafter{\romannumeral6}}}&\multirowcell{4}{
    % Experiment \hypertarget{6}{
    $\lambda$
    % \uppercase\expandafter{\romannumeral6}}
    }&1 & 76.3 & 123.4 \\
            &&2 & 75.8 & 121.8 \\
            &&3 (B/L) & 75.7 & 123.4 \\
            &&4 & 75.7 & 123.4 \\
    \end{tabular}
    \begin{tabular}{m{0.7cm}<{\centering}|m{0.8cm}<{\centering}
    m{1.13cm}<{\centering} 
m{1.05cm}<{\centering}
m{1.05cm}<{\centering}
m{1.05cm}<{\centering}
}
\toprule[1pt]
&\multicolumn{2}{c}{\multirow{2}{*}{Method}}&  \multicolumn{3}{c}{mIoU(\%)}\\\cline{4-6}
&&& Stage 2\rule{0pt}{9pt} & Stage 3& Stage 4\\\midrule[1pt]
\multirowcell{4}{\hypertarget{7}{\uppercase\expandafter{\romannumeral7}}}&\multirowcell{4}{
% Experiment \hypertarget{7}{
$\delta$
% \uppercase\expandafter{\romannumeral7}}
}&-2	&74.9	&75.7	&75.2\\
&&-1	&75.6	&76	&75.6\\
&&0 (B/L)	&76.3	&76.3	&76.3\\
&&1	&76.4	&76.1	&76.3\\
\bottomrule[1.5pt]
\end{tabular}
    \caption{Ablation experiments on the network structure design on Cityscapes. The baseline (B/L) method for all ablation experiments is the DWRSeg-L model, and all results are obtained under the input size of 1536 $\times$ 768. SS denotes a single receptive field toward a single feature map’, MS denotes multiple receptive fields toward a single feature map’, $\alpha$ denotes the ratio of output channel numbers in three branches of the DWR module, $\beta$ denotes the ratio of channel expansion of Region Residualization in the DWR module, NL denotes nonlinearity, PWC denotes point-wise convolution in the SIR module, $\lambda$ denotes the ratio of the output channel number to the input channel number of beginning convolution in the SIR module and $\delta$ denotes the block number offset to the baseline.
}
   
    \label{tab:ab}
\end{table}

\textbf{Single Receptive Field toward a Single Feature Map.}
\Cref{tab:ab} Expt. \hyperlink{1}{\uppercase\expandafter{\romannumeral1}} illustrates the results of a 'single receptive field toward a single feature map' ('SS') and 'multiple receptive fields toward single feature map' ('MS') in the DWR module.
The results show that 'SS' vastly improves the output.

\textbf{Ratios of Different Receptive Fields in the DWR module.}
\Cref{tab:ab} Expt. \hyperlink{2}{\uppercase\expandafter{\romannumeral2}} investigates the effects of different ratios $\alpha$ of the output channel numbers of the three branches (d-1:d-3:d-5) in the DWR module. When we adjust $\alpha$, the best effect is obtained when the ratio is 2:1:1.

\textbf{Channel Expansion of Region-Residualization in the DWR Module.} 
\Cref{tab:ab} Expt. \hyperlink{3}{\uppercase\expandafter{\romannumeral3}} shows the results of setting a different ratio $\beta$ of the output channel number to the input channel number of the Region Residualization in the DWR module. The best output is achieved when $\beta$ is set to 1.5.

\textbf{Nonlinearity in the DWR Module.}
\Cref{tab:ab} Expt. \hyperlink{4}{\uppercase\expandafter{\romannumeral4}} illustrates the impact of different degrees of nonlinearity in the DWR module. Sitch 1 represents Region Residualization without the ReLU, Sitch 2 represents Region Residualization without BN, Sitch 3 represents Semantic Residualization without BN, Sitch 4 represents adding an ReLU after BN in Semantic Residualization and Sitch 5 represents adding a BN after point-wise convolution.
The results show that when the nonlinearity is lower, the effect decreases considerably; however, increasing the nonlinearity does not improve the output.

\textbf{Depth-wise Convolution in the SIR Module.}
As shown in \Cref{tab:ab} Expt. \hyperlink{5}{\uppercase\expandafter{\romannumeral5}}, the improvement obtained by adding a depth-wise convolution with a BN layer ('w') is less than that obtained by increasing the number of modules ('w/o') in the SIR module.

\textbf{Channel Expansion in the SIR Module.} 
\Cref{tab:ab} Expt. \hyperlink{6}{\uppercase\expandafter{\romannumeral6}} shows the effects of expanding the channel number of the beginning convolution to $\lambda$ times in the SIR module.
The best result is obtained when the number of channels expands to three times.

\textbf{Block Number.} 
We adjust the number of modules and present the results in \Cref{tab:ab} Expt. \hyperlink{7}{\uppercase\expandafter{\romannumeral7}}. $\delta$ represents the module number offset to the baseline. Each column shows the results obtained when the number of modules in the corresponding stage changes according to $\delta$, while the number of modules in other stages remains unchanged.
The benefits of more blocks appreciably decrease, and a deeper network is detrimental to parallel calculation and FPS.

\subsection{Comparison with State-of-the-Art Methods}
In this section, we compare our models with other state-of-the-art methods on two datasets: Cityscapes and CamVid.

\textbf{Results on Cityscapes.} \Cref{tab:city} presents the segmentation accuracy, inference speed and model parameters of our proposed method on the Cityscapes test set. Following the previous methods, we use the training set and validation set to train our models before submitting them to the Cityscapes online server. We use 50 and 75 after the method name to represent the input size 512 $\times$ 1024 and 768 $\times$ 1536, respectively.
Our DWRSeg-L75 achieves the best mIoU of 76.3\% at a speed of 123.4 FPS, and our DWRSeg-B50 achieves the fastest speed of 319.5 FPS with an mIoU of 72.7\%. Compared with the second-best model, the STDC, which uses pretraining and auxiliary segmentation, our models produce a \textbf{state-of-the-art} trade-off between accuracy and speed without pretraining or resorting to any training trick. Moreover, our models are much lighter than STDC, with only a quarter of the STDC parameters.

\begin{table}
 \renewcommand{\arraystretch}{0.9}
    \begin{tabular}{l|c|c|c|c}
        \toprule[1.5pt]
        Model &\begin{tabular}[c]{@{}c@{}}Input\\Ratio\end{tabular}&\begin{tabular}[c]{@{}c@{}}mIoU\\(\%)\end{tabular}&FPS&\begin{tabular}[c]{@{}c@{}}Params\\(M)\end{tabular}\\ \midrule[1pt]
        ENet~\cite{paszke2016enet} & 0.5&58.3 & 76.9 &\textbf{0.37}\\
        ICNet$\dagger$~\cite{zhao2018icnet}& 1.0&69.5 & 30.3 &26.5\\
        DABNet~\cite{li2019dabnet} & 1.0&70.1 & 27.7 &0.76\\
        DFANet B$\dagger$~\cite{li2019dfanet}& 1.0&67.1 & 120 &4.8\\
        DFANet A$\dagger$~\cite{li2019dfanet}& 1.0&71.3 & 100 &7.8\\
        %BiSeNetv1(Xception39)$\dagger$ & 0.75&68.4 & 105.8 &5.8\\
        BiSeNetV2~\cite{yu2021bisenet} & 0.5&72.6 & 156 &2.33\\
        % BiSeNetV2-L~\cite{yu2021bisenet} & 0.5&75.3 & 47.3 &18.41\\
        DF1-Seg~\cite{li2019partial} & 1.0&73.0 & 80 &8.55\\
        DF2-Seg~\cite{li2019partial}& 1.0&74.8 & 55 &8.55\\
        SFNet(DF1)~\cite{li2020semantic} & 1.0&74.5 & 121 &9.03\\
        STDC1-Seg50$\dagger$~\cite{fan2021rethinking} & 0.5&71.9 & 250.4 &9.97\\
         STDC2-Seg50$\dagger$~\cite{fan2021rethinking} & 0.5&73.4 & 188.6 &14.0\\
        STDC1-Seg75$\dagger$~\cite{fan2021rethinking} & 0.75&75.3 & 126.7 &9.97\\
         STDC2-Seg75$\dagger$~\cite{fan2021rethinking} & 0.75&\textbf{76.8} & 97.0 &14.0 \\
        \midrule[1pt]
        DWRSeg-B50 & 0.5&72.7 & \textbf{319.5} &2.54\\
        DWRSeg-L50 & 0.5&73.1 & 256.2 &3.53\\
        DWRSeg-B75 & 0.75&75.6 & 151.7 &2.54\\
        DWRSeg-L75 & 0.75&76.3 & 123.4 &3.53\\
        \bottomrule[1.5pt]
    \end{tabular}
    \caption{Comparisons with other state-of-the-art methods on Cityscapes. The base resolution is 1024$\times$2048. $\dagger$ means that the model is pretrained on ImageNet~\cite{deng2009imagenet}.}
\label{tab:city}
\end{table}

\textbf{Results on CamVid.} \Cref{tab:camvid} shows the comparison results with other methods.  DWRSeg-B achieves 76.5\% mIoU at 237.2 FPS and DWRSeg-L achieves 77.5\% mIoU at 189.2 FPS. This further demonstrates the superior capability of our method.

\begin{table}
\centering
 \renewcommand{\arraystretch}{0.95}
    \begin{tabular}{m{2.5cm} 
            m{1.5cm}<{\centering}
            m{1.5cm}<{\centering}}
        \toprule[1.5pt]
        Model &mIoU(\%)&FPS\\ \midrule[1pt]
        % ENet &51.3 & 61.2 \\
        ICNet$\dagger$~\cite{zhao2018icnet}&67.1 & 34.5 \\
        DFANet A$\dagger$~\cite{li2019dfanet}&64.7 & 120 \\
        %BiSeNetv1 &65.6 & 175 \\
        
        %BiSeNetV2-L &73.2 & 32.7 \\
        STDC1$\dagger$~\cite{fan2021rethinking}&73.0 & 197.6 \\
        STDC2$\dagger$~\cite{fan2021rethinking}&73.9 & 152.2 \\
        BiSeNetV2*~\cite{yu2021bisenet}&76.7 & 124.5 \\
        \midrule[1pt]
        DWRSeg-B* &76.5 & 237.2 \\
        DWRSeg-L* &77.5 & 189.2 \\
        
        \bottomrule[1.5pt]
    \end{tabular}
    \caption{Comparisons with other state-of-the-art methods on CamVid. $\dagger$ means the models are pretrained on ImageNet; * means the models are pretrained on Cityscapes.}
\label{tab:camvid}
\end{table}

\section{Conclusion}
In this paper, to improve the efficiency of drawing multi-scale contextual information in real-time semantic segmentation tasks, we propose a two-step residual feature extraction method, Region Residualization – Semantic Residualization. Based on this idea, we design the DWR module and SIR module with elaborated receptive fields for the upper and lower stages of the network, respectively. Using these modules, we introduce a DWRSeg network. Extensive ablation experiments indicate the effectiveness of our proposed DWRSeg
networks, and the comparison experiments with other networks demonstrate that our networks reach a state-of-the-art trade-off between accuracy and speed. Moreover, our networks have a lighter weight. We hope that the idea and structure of DRWSeg foster further research in semantic segmentation.

\textbf{Acknowledgment.} This research is supported by Beijing Municipal Natural Science Foundation (No. L191020).
{\small
\bibliographystyle{ieee_fullname}
\bibliography{egbib}
}

\end{document}